\documentclass[conference]{IEEEtran}
\IEEEoverridecommandlockouts

\usepackage{cite}
\usepackage{amsmath,amssymb,amsfonts}
\usepackage{graphicx}
\usepackage{subcaption}
\usepackage{textcomp}
\usepackage{xcolor}
\usepackage{pifont}

\usepackage{algorithm}
\usepackage{algpseudocode}

\usepackage{multirow}

\usepackage[T1]{fontenc}
\usepackage[utf8]{inputenc}

\makeatletter
\newcommand{\linebreakand}{%
  \end{@IEEEauthorhalign}
  \hfill\mbox{}\par
  \mbox{}\hfill\begin{@IEEEauthorhalign}
}
\makeatother

\def\BibTeX{{\rm B\kern-.05em{\sc i\kern-.025em b}\kern-.08em
    T\kern-.1667em\lower.7ex\hbox{E}\kern-.125emX}}
\begin{document}

\title{FedQP: Towards Accurate Federated Learning using Quadratic Programming Guided Mutation}

\author{
	\IEEEauthorblockN{
		Jiawen Weng\IEEEauthorrefmark{2}, 
		Zeke Xia\IEEEauthorrefmark{2}, 
		Ran Li\IEEEauthorrefmark{3}, 
		Ming Hu\IEEEauthorrefmark{4},
		Mingsong Chen\thanks{* Mingsong Chen (mschen@sei.ecnu.edu.cn) is the corresponding author. DOI reference number: 10.18293/SEKE2024-108}\IEEEauthorrefmark{2}} 
	\IEEEauthorblockA{\IEEEauthorrefmark{2}Software/Hardware Co-design Engineering Research Center, Ministry of Education, \\East China Normal University, Shanghai, China}
	\IEEEauthorblockA{\IEEEauthorrefmark{3}School of Software, Nanjing University of Information Science and Technology, Nanjing, China}
	\IEEEauthorblockA{\IEEEauthorrefmark{4}Research Lab for Intelligent Software Engineering, Singapore Management University, Singapore} 
} 


\maketitle

\begin{abstract}
Due to the advantages of privacy-preserving, Federated Learning (FL) is widely used in distributed machine learning systems.
However, existing FL methods suffer from low-inference performance caused by data heterogeneity.
Specifically, due to heterogeneous data, the optimization directions of different local models vary greatly, making it difficult for the traditional FL method to get a generalized global model that performs well on all clients.
As one of the state-of-the-art FL methods, the mutation-based FL method attempts to adopt a stochastic mutation strategy to guide the model training towards a well-generalized area (i.e., flat area in the loss landscape).
Specifically, mutation allows the model to shift within the solution space, providing an opportunity to escape areas with poor generalization (i.e., sharp area).
However, the stochastic mutation strategy easily results in diverse optimal directions of mutated models, which limits the performance of the existing mutation-based FL method.
To achieve higher performance, this paper proposes a novel mutation-based FL approach named FedQP, utilizing a quadratic programming strategy to regulate the mutation directions wisely.
By biasing the model mutation towards the direction of gradient update rather than traditional random mutation, FedQP can effectively guide the model to optimize towards a well-generalized area (i.e., flat area).
Experiments on multiple well-known datasets show that our quadratic programming-guided mutation strategy effectively improves the inference accuracy of the global model in various heterogeneous data scenarios.


\end{abstract}


\section{Introduction}
Federated Learning (FL)~\cite{fedavg,FedProx,hu2023gitfl,huang2023rethinking, xia2024cabafl,hu2024icde,wang2023aocc,yan2023have,hu2024kdd,yang2023protect} is a distributed machine learning framework that enables multiple data owners to collaboratively train a model without exchanging their original data.
Due to its advantage of protection of privacy, FL has been used in various security-critical fields, including healthcare~\cite{nguyen2022federated}, AIoT~\cite{jia2024dac,zhang2020efficient,hu2023aiotml,song2021tii}, and autonomous driving~\cite{auto_drive_ijcnn}. 
Most FL conducts a central server-based architecture, which includes a cloud server and massive local clients.
During each training round, clients train their local models on their raw data and then upload the trained models to the cloud server. The cloud server aggregates these models to create a global model, which is then distributed back to all clients.
This approach allows FL to reduce the risk of privacy leakage significantly.

Although FL exhibits effectiveness for privacy protection, its performance is still constrained by data heterogeneity problems.
Specifically, since data distributions among clients are typically non-Independent and Identically Distributed (non-IID)~\cite{non-iid}, resulting in different optimization directions in each client, thereby leading to a large number of conflict gradients among local models, known as ``gradient divergence'' problem~\cite{scaff}.
Aggregating these conflicting gradients inevitably degrades the performance of the global model.

To overcome these challenges, several FL optimization techniques have been proposed, such as global variable adjustment~\cite{scaff}, client grouping strategies~\cite{cs}, and knowledge distillation technology~\cite{fedntd}. 
These methods can improve inference performance to some extent, but they still have shortcomings, such as relying on an additional public dataset, being incompatible with secure aggregation, or resulting in additional non-negligible communication costs.

To effectively alleviate performance degradation caused by heterogeneous data without the need for additional data or communication overhead, the mutation-based FL method~\cite{hu2024fedmut} has been proposed.
It has been observed that well-generalized solutions tend to be located in flat valleys instead of sharp ravines~\cite{hochreiter1994simplifying}. So Mutation-based FL modifies the global model based on gradients to create several intermediate models and then sends them to various clients for local training. 
In this way, each intermediate model can be shifted within the solution space and easily jump out of local optimal solutions, i.e., sharp area.
However, the existing mutation-based FL method, i.e., FedMut, only performs a coarse-grained mutation strategy.
Specifically, FedMut randomly adds or minus the gradients for each layer of the global model to produce intermediate models.
In this way, the mutated intermediate models are located in a circular area.
However, too diverse mutation directions lead to intermediate models optimized towards diverse directions, leading to gradient conflict, which finally results in smaller aggregation gradients.
Since mutation is based on gradients, too small gradients limit the range of mutation, making it more difficult for the mutated model to jump out of sharp areas.
Therefore, {\it how to wisely limit the range of mutation directions to effectively guide mutated models towards a more flat area is an important problem in mutation-based FL.}

To tackle the issue mentioned above, we propose a new approach called FedQP, which employs a quadratic programming strategy to regularize the range of mutations. Specifically, FedQP aims to guide intermediate models to mutate into a fan-shaped area towards the direction of aggregated gradients, rather than a circular area.
In this way, FedQP can achieve a biased mutation, alleviating the limited mutation range caused by gradient conflicts.

In summary, this paper has the following three contributions:

\begin{itemize}
    \item We propose a quadratic programming-guided mutation strategy, which uses quadratic programming to regularize the range of mutation to improve the effectiveness of model training.

    \item We implement a novel mutation-based FL framework, named FedQP, based on our proposed quadratic programming-guided mutation strategy.

    \item We carried out experiments on well-known datasets using various models and showed the effectiveness of our approach for both IID and non-IID cases.
\end{itemize}

\section{Preliminary and Related Work}

\subsection{Preliminary}

In a standard FL scenario, we assume that there are $ N $ participants, each with its own private dataset $ \mathcal{D}_i $ of size $ n_i $. These participants work together to train a global model without sending their datasets to a cloud server or sharing them with other participants.

The global goal is to minimize the weighted average of all participants' local objective functions. The specific formula is as follows:

\begin{equation}
\min_{\mathbf{w}} \quad f(\mathbf{w}) = \sum_{i=1}^N \frac{n_i}{n} F_i(\mathbf{w}),
\end{equation}
where $ \mathbf{w} $ represents the parameters of the global model, $ F_i(\mathbf{w}) $ is the loss function for client $i$, representing the loss of the model with $ \mathbf{w} $ as parameters on $\mathcal{D}_i$ data. The total amount of data across all clients is given by $ n = \sum_{i=1}^N n_i $.

\subsection{Related Work}
Although the classic FedAvg provides a concise but effective paradigm. However, the generalization of FedAvg in non-IID scenarios is severely limited. There are many works that focus on improving the performance of the global model. For instance, FedProx\cite{FedProx} adds a penalty term to the local loss function through regularization to constrain the update direction of the local model. CluSamp\cite{cs} clusters based on the cosine similarity between client gradient updates and selects devices from different clusters for local training. There is also a class of methods to improve the accuracy of global models through knowledge distillation. For Example, FedNTD\cite{fedntd} alleviates the weight divergence problem through knowledge distillation to alleviate the data heterogeneity problem. FedGen\cite{fedgen} also uses knowledge distillation but uses a generator to generate distilled data so that no proxy dataset is required.  However, the above methods still lack good generalization of the global model.
To the best of our knowledge, our method is the first to employ quadratic programming to guide the model mutating process.

\section{Approach}
\subsection{Motivation}

Inspired by mutation-based FL, it is feasible to generate mutated models to enable the global model to escape the sharp solution region. Our goal is to guide the model to generate biased mutations through a fine-grained mutation strategy. Intuitively, if the mutation range of a model is within a circular area, the aggregated results of the generated mutated models will have a higher probability of remaining in this area, and effective exploration of the parameter space cannot be achieved. On the contrary, if the mutation range of a model is a fan-shaped area, the aggregated results of the mutated models will be more likely to escape the current area. Based on this intuition, we constrain the range of mutation generation and produce several mutated models for local training in the fan-shaped area guided by the gradient, allowing the model to conduct more concentrated exploration in the direction of gradient descent. In theory, this design can guide the model to accelerate convergence to the high-performance area.



\begin{figure}[h]
    \centering
    \begin{subfigure}[b]{0.25\textwidth}
        \includegraphics[width=\textwidth]{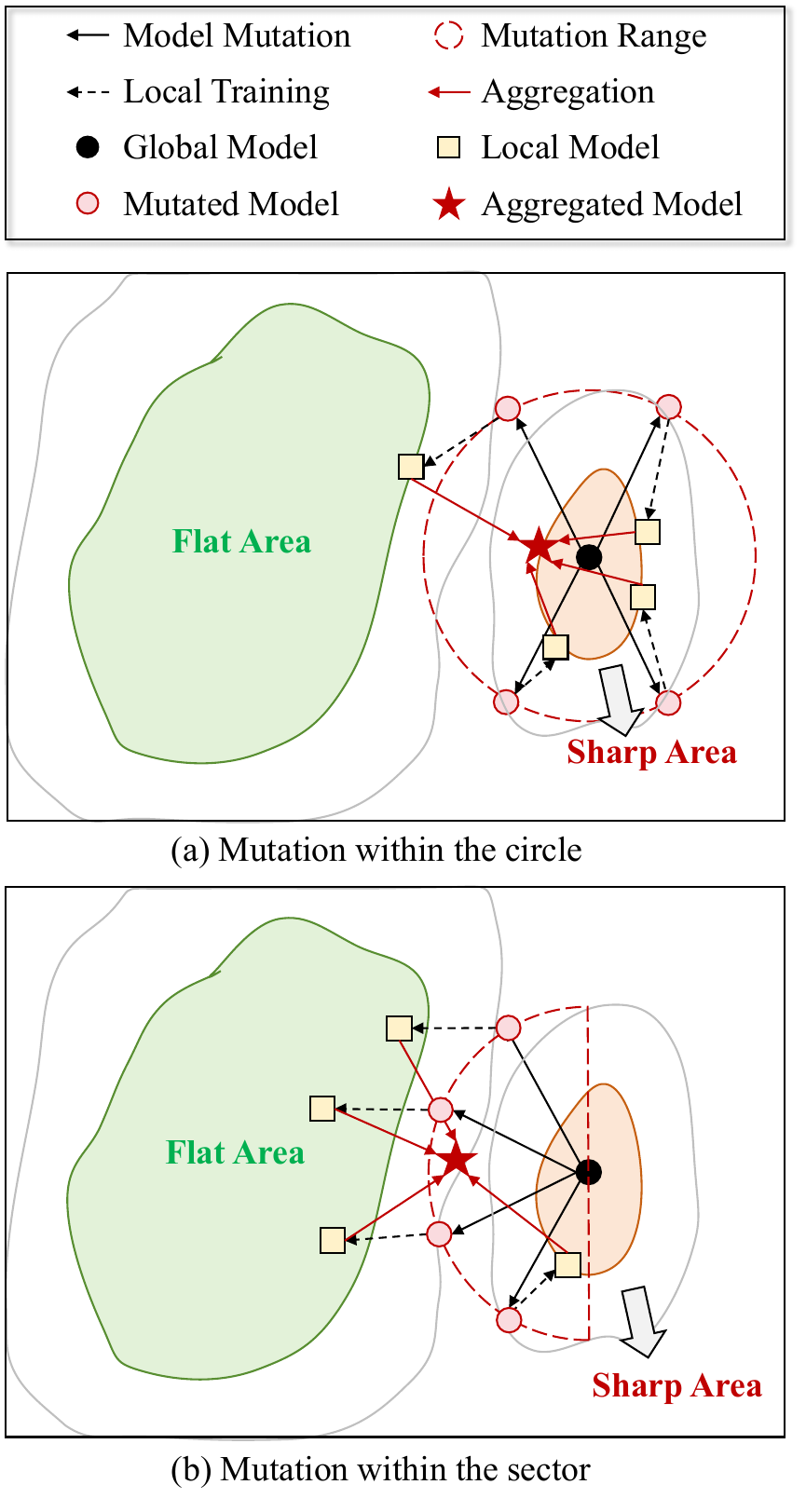}
    \end{subfigure}
    \caption{Motivation of mutation range.}
    \label{fig:motivation-fig}
\end{figure}

\begin{figure*}[h]
    \centering
    \begin{subfigure}[b]{0.6\textwidth}
        \includegraphics[width=\textwidth]{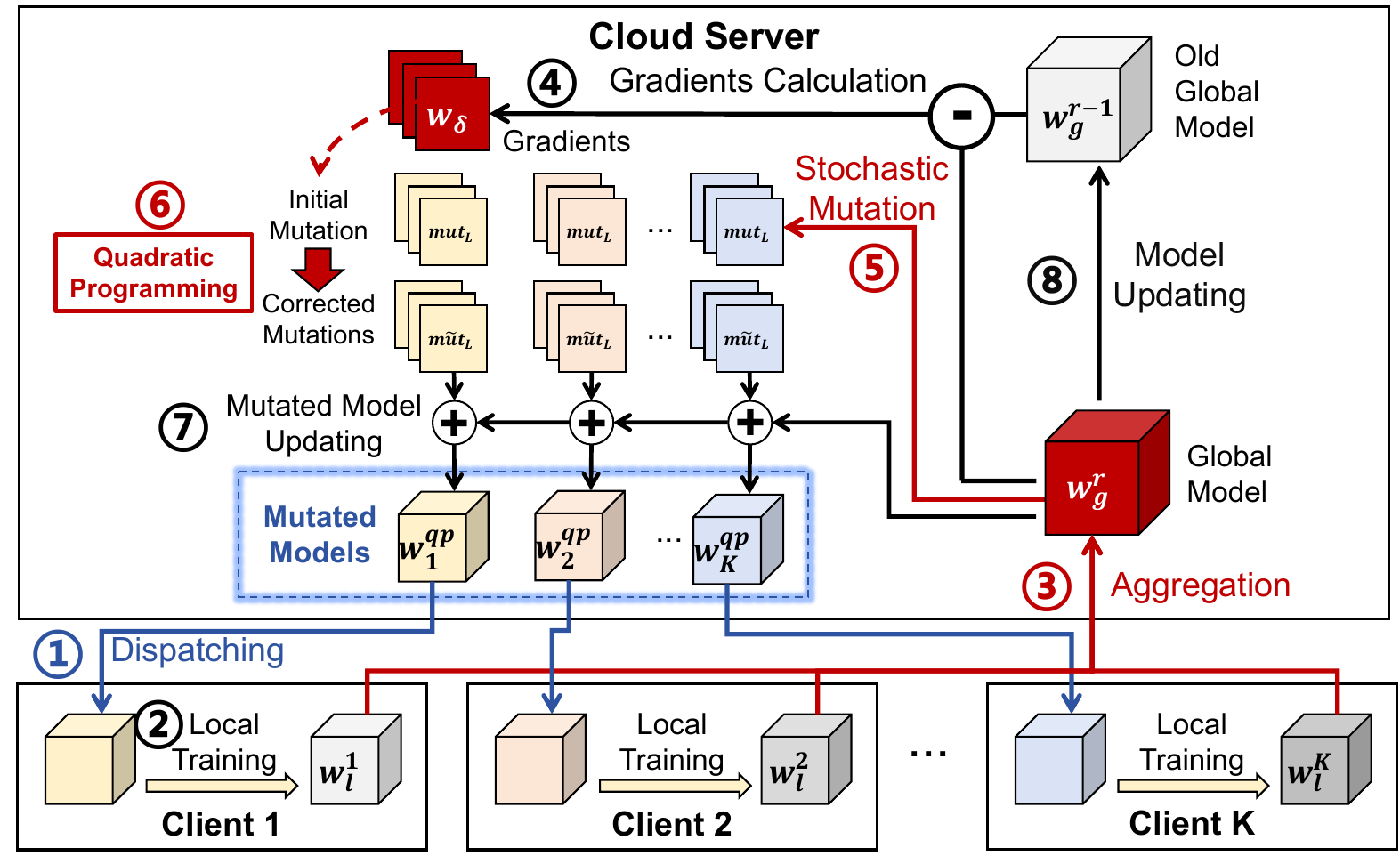}
    \end{subfigure}
    \vspace{-0.05 in}
    \caption{Framework and workflow of our approach.}
    \vspace{-0.25 in}
    \label{framework}
\end{figure*}

Figure \ref{fig:motivation-fig} shows the motivation of our method. In these two sub-figures, we can see that the green area on the left side of the figure represents a flat solution in parameter space, while the red area on the right side represents a sharp solution. In Figure \ref{fig:motivation-fig}(a), the model mutation of FedMut is shown, and Figure \ref{fig:motivation-fig}(b) presents the fan-shaped mutation after our optimization. In the figure, the black dots represent the original global model. The black arrows represent the mutation routes of the original global model. The red dots represent four models mutated from the global model, and the red dashed line represents the mutation range. After the local training process indicated by the black dashed arrows, we can get the local model represented by the yellow box. Finally, after the average aggregation of the red arrow, the red five-pointed star is obtained as the aggregated global model.

Figure \ref{fig:motivation-fig}(a) is the uniform random mutation without directionality. It may cause three mutated models to still optimize towards the sharp area so that the global model obtained by the final aggregation has the problem of moving slowly or even not moving towards the flat area. In Figure 1(b), we associate the fan-shaped opening with the direction of the global gradient. Therefore, all mutations in this round are along the direction of global gradient descent, more mutated models can move towards the flat area after local training, and then the aggregated global model will also accelerate the convergence to the flat area.

\subsection{Overview}
Figure \ref{framework} demonstrates the framework of our FedQP. In our framework, there is a server and $N$ local devices. The server maintains a global model and $K$ mutated models. Unlike traditional federated learning, we do not dispatch the global model to the device for local training but dispatch the mutated model. As shown in Figure \ref{framework}, our method comprises the following eight steps. \textbf{Step 1 (Dispatching):} The server selects $K$ devices and sends $K$ mutated models to these devices by random. \textbf{Step 2 (Local Training):} Each device trains the received model using its local data and uploads the trained model to the server. \textbf{Step 3 (Aggregation):} After receiving the models from all devices, the server aggregates these models to obtain a new global model. \textbf{Step 4 (Gradients Calculation):} The server subtracts the global model of this round from that of the previous round to obtain the global gradient. \textbf{Step 5 (Stochastic Mutation):} The server applies random perturbations to the global model to generate the initial mutations. \textbf{Step 6 (Quadratic Programming):} The server uses the global gradient to perform quadratic programming on the initial mutations to obtain the corrected mutations. \textbf{Step 7 (Mutated Model Updating):} The server adds the global model to the corrected mutations to update the mutated models. \textbf{Step 8 (Model Updating):} The server replaces the global model of the previous round with the global model of the current round.

\subsection{Implementation}

\begin{algorithm}[h]
	\caption{Implementation of our method}
        \label{impl}
	\begin{algorithmic}[1]
		\State \textbf{Input:} 1) $R$, \# of FL rounds; 2) $D$, total involved devices; 3) $K$, \# of selected devices in one FL round; 4) $p$, QP Probability
		\State \textbf{Output:} $w_g$, the global model.
		\State \textbf{FedQP}($R,C,K$)
            \State initialize the global model $w_g^0$\\
            $w_1^{qp},...,w_K^{qp}\leftarrow w_g^0$
            \For{$r=1,...,R$}
            \State $S\leftarrow$ Randomly select $K$ clients from $C$
            \For{$i=1,...,K$}
            \State $w_l^i\leftarrow LocalTraining(w^{qp}_i,S[i])$
            \EndFor
            \State $w_g^r\leftarrow Aggregation(w_l^1,...,w_l^K)$
            \State $w_{\delta}\leftarrow w_g^r-w_g^{r-1}$
		\For{$j=1,...,K$}
		\For{each layer $L$ in $w_g^r$}
		\State $mut_L\leftarrow$ Randomly generate mutation of L
            \State $\tilde{mut}_L\leftarrow mut_L$
		\State $\tilde{mut}_L \gets$ \Call{QPMut}{$w_{\delta}.L, mut_L,p$}
		\EndFor
		\State $w_j^{qp}\leftarrow w_l^j+\tilde{mut}$
		\EndFor
            \EndFor
            \State \textbf{Return} $w_g^r$
	\end{algorithmic}
\end{algorithm}
\vspace{-0.1 in}

Algorithm \ref{impl} details the realization of our method. In Line 4, the global model $w_g^0$ is initialized, and in Line 5, all mutated models are initialized. Lines 6-20 describe the overall FL training process. Line 7 indicates the server randomly selects $K$ devices from all devices $D$ to participate in the current round of FL training. In lines 8-10, the server sends the $K$ mutated models to the selected $K$ devices for local training. After the device completes local training, it uploads the updated models to the server. Line 11 indicates the server aggregates the received local models and obtains a new global model. In Line 12, the server subtracts the old and new global models to get the gradient of the current round. Lines 13-20 present our quadratic programming-guided mutation process. The for loop in line 13 means that we generate $K$ new mutated models using the global model. First, we randomly mutate each layer of the model (Line 15). Then, the server will have probability $p$ to perform quadratic programming on the generated random mutation (Line 17). Finally, the server adds the global model to the mutation to get a new mutated model (Line 19).

\subsection{QP-based model mutation(QPMut($\cdot$))}
For ease of writing, we omit $L$ in the following sections.
In our method, we want to constrain the circular mutation range in FedMut to a fan-shaped area guided by the global gradient. Therefore, we perform quadratic programming on the mutants generated by the mutation, requiring that the Euclidean distance between the corrected mutant and the original mutant is the shortest and the angle with the global gradient direction is an acute angle. We define the objective function and constraints of the quadratic programming as follows:
\begin{equation}\label{eq1}
\min_{\tilde{mut}} \  \| \tilde{mut} - mut\|^2 , s.t. \ \left \langle \tilde{mut} , w_{\delta} \right \rangle \ge 0,
\end{equation}
where \( \tilde{mut} \) represents the mutation corrected according to the gradient direction, and $mut$ represents the original mutation.

This form of objective function ensures that the distance between the adjusted mutation and the original mutation is as small as possible, that is, to find the solution closest to the original mutation. The constraints ensure that the adjusted mutation is consistent with the direction of the global gradient, which helps to ensure that the mutation operation does not deviate from the correct direction of optimization and enhances the algorithm's convergence.

After expanding and simplifying the above equation and discarding the constant term, we rewrite equation \ref{eq1}  as follows:
\begin{equation}\label{eq2}
\min_{\tilde{mut}} \left (   \frac{1}{2} \tilde{mut}^\top \tilde{mut} - mut^\top \tilde{mut}  \right ), s.t. w_{\delta}^\top \tilde{mut} \geq 0.
\end{equation}

However, directly solving this problem may involve expensive computational costs when dealing with optimization problems with high-dimensional parameter spaces. To this end, we consider leveraging the dual form of quadratic programming to simplify the solution process.

By introducing the Lagrange multiplier, we can calculate:
\begin{equation}\label{eq3}
\tilde{mut} = mut + w_{\delta}\lambda. 
\end{equation}

We get the dual problem of the equation \ref{eq2} after simplification:
\begin{equation}\label{eq4}
 \min_\lambda \left( \frac{1}{2} \lambda^\top w_{\delta}^\top w_{\delta}\lambda + mut^\top w_{\delta}\lambda \right), s.t. 
\lambda \geq 0.   
\end{equation}

Observing formula \ref{eq3}, we can find that the output $\tilde{mut}$ of the quadratic programming is essentially a conical combination of the gradient $w_{\delta}$ and $mut$. Intuitively, this suggests that our mutation method can adaptively exploit global knowledge, helping us accelerate training and improve generalization.

\section{Experiment}
\begin{table*}[h]
\centering
\caption{Comparison of test accuracy.}
\label{tab:accuracy-comparison}
\resizebox{\textwidth}{!}{%
\begin{tabular}{|c|c|c|cccccc|}
\hline
\multirow{2}{*}{DataSet} & \multirow{2}{*}{Model} & \multirow{2}{*}{Hetero.Settings} & \multicolumn{6}{c|}{Test Accuracy (\%)} \\ \cline{4-9} 
 &  &  & \multicolumn{1}{c|}{FedAvg} & \multicolumn{1}{c|}{CluSamp} & \multicolumn{1}{c|}{FedNTD} & \multicolumn{1}{c|}{FedGen} & \multicolumn{1}{c|}{FedMut} & Ours \\ \hline \hline
\multirow{12}{*}{CIFAR-10} & \multirow{4}{*}{CNN} & $\beta=0.1$ & \multicolumn{1}{c|}{$47.10\pm3.07$} & \multicolumn{1}{c|}{$47.32\pm2.28$} & \multicolumn{1}{c|}{$50.92\pm0.98$} & \multicolumn{1}{c|}{$47.07\pm2.65$} & \multicolumn{1}{c|}{$52.22\pm1.32$} & $\textbf{53.04}\pm\textbf{0.87}$ \\ \cline{3-9} 
 &  & $\beta=0.5$ & \multicolumn{1}{c|}{$53.48\pm0.78$} & \multicolumn{1}{c|}{$53.38\pm1.03$} & \multicolumn{1}{c|}{$55.26\pm0.33$} & \multicolumn{1}{c|}{$53.96\pm1.30$} & \multicolumn{1}{c|}{$55.51\pm0.72$} & $\textbf{55.66}\pm\textbf{0.28}$ \\ \cline{3-9} 
 &  & $\beta=1.0$ & \multicolumn{1}{c|}{$57.11\pm0.66$} & \multicolumn{1}{c|}{$57.28\pm0.87$} & \multicolumn{1}{c|}{$57.58\pm0.29$} & \multicolumn{1}{c|}{$52.22\pm1.32$} & \multicolumn{1}{c|}{$58.04\pm0.25$} & $\textbf{61.06}\pm\textbf{0.37}$ \\ \cline{3-9} 
 &  & IID & \multicolumn{1}{c|}{$57.46\pm0.22$} & \multicolumn{1}{c|}{$57.88\pm0.21$} & \multicolumn{1}{c|}{$58.88\pm0.15$} & \multicolumn{1}{c|}{$57.03\pm1.32$} & \multicolumn{1}{c|}{$60.00\pm0.18$} & $\textbf{61.15}\pm\textbf{0.16}$ \\ \cline{2-9} 
 & \multirow{4}{*}{ResNet-18} & $\beta=0.1$ & \multicolumn{1}{c|}{$44.12\pm3.11$} & \multicolumn{1}{c|}{$43.09\pm3.22$} & \multicolumn{1}{c|}{$51.17\pm2.08$} & \multicolumn{1}{c|}{$44.35\pm1.87$} & \multicolumn{1}{c|}{$53.60\pm4.69$} & $\textbf{55.66}\pm\textbf{0.28}$ \\ \cline{3-9} 
 &  & $\beta=0.5$ & \multicolumn{1}{c|}{$61.72\pm0.58$} & \multicolumn{1}{c|}{$61.35\pm0.48$} & \multicolumn{1}{c|}{$64.96\pm0.24$} & \multicolumn{1}{c|}{$60.92\pm0.69$} & \multicolumn{1}{c|}{$68.60\pm0.16$} & $\textbf{74.09}\pm\textbf{0.15}$ \\ \cline{3-9} 
 &  & $\beta=1.0$ & \multicolumn{1}{c|}{$64.28\pm0.35$} & \multicolumn{1}{c|}{$65.83\pm0.22$} & \multicolumn{1}{c|}{$68.10\pm0.12$} & \multicolumn{1}{c|}{$65.80\pm0.34$} & \multicolumn{1}{c|}{$69.95\pm0.23$} & $\textbf{75.75}\pm\textbf{0.22}$ \\ \cline{3-9} 
 &  & IID & \multicolumn{1}{c|}{$65.42\pm0.18$} & \multicolumn{1}{c|}{$65.29\pm0.11$} & \multicolumn{1}{c|}{$69.10\pm0.13$} & \multicolumn{1}{c|}{$63.88\pm0.42$} & \multicolumn{1}{c|}{$66.58\pm0.13$} & $\textbf{71.73}\pm\textbf{0.07}$ \\ \cline{2-9} 
 & \multirow{4}{*}{VGG-16} & $\beta=0.1$ & \multicolumn{1}{c|}{$56.20\pm8.68$} & \multicolumn{1}{c|}{$60.52\pm7.62$} & \multicolumn{1}{c|}{$64.95\pm0.68$} & \multicolumn{1}{c|}{$64.05\pm3.62$} & \multicolumn{1}{c|}{$64.60\pm3.47$} & $\textbf{69.88}\pm\textbf{3.43}$ \\ \cline{3-9} 
 &  & $\beta=0.5$ & \multicolumn{1}{c|}{$76.97\pm0.85$} & \multicolumn{1}{c|}{$75.86\pm1.58$} & \multicolumn{1}{c|}{$80.14\pm0.41$} & \multicolumn{1}{c|}{$78.00\pm0.77$} & \multicolumn{1}{c|}{$80.86\pm0.23$} & $\textbf{82.10}\pm\textbf{0.33}$ \\ \cline{3-9} 
 &  & $\beta=1.0$ & \multicolumn{1}{c|}{$79.21\pm0.62$} & \multicolumn{1}{c|}{$79.45\pm0.88$} & \multicolumn{1}{c|}{$81.74\pm0.55$} & \multicolumn{1}{c|}{$79.01\pm0.65$} & \multicolumn{1}{c|}{$81.08\pm0.21$} & $\textbf{83.27}\pm\textbf{0.12}$ \\ \cline{3-9} 
 &  & IID & \multicolumn{1}{c|}{$80.82\pm0.08$} & \multicolumn{1}{c|}{$80.24\pm0.17$} & \multicolumn{1}{c|}{$82.36\pm0.11$} & \multicolumn{1}{c|}{$80.01\pm0.08$} & \multicolumn{1}{c|}{$81.73\pm0.08$} & $\textbf{83.58}\pm\textbf{0.03}$ \\ \hline \hline
\multirow{12}{*}{CIFAR-100} & \multirow{4}{*}{CNN} & $\beta=0.1$ & \multicolumn{1}{c|}{$29.02\pm1.03$} & \multicolumn{1}{c|}{$29.51\pm0.61$} & \multicolumn{1}{c|}{$30.34\pm0.43$} & \multicolumn{1}{c|}{$28.32\pm1.08$} & \multicolumn{1}{c|}{$30.45\pm0.34$} & $\textbf{30.82}\pm\textbf{1.00}$ \\ \cline{3-9} 
 &  & $\beta=0.5$ & \multicolumn{1}{c|}{$31.33\pm0.86$} & \multicolumn{1}{c|}{$32.28\pm0.42$} & \multicolumn{1}{c|}{$34.28\pm0.25$} & \multicolumn{1}{c|}{$32.52\pm0.54$} & \multicolumn{1}{c|}{$35.39\pm0.31$} & $\textbf{35.98}\pm\textbf{0.32}$ \\ \cline{3-9} 
 &  & $\beta=1.0$ & \multicolumn{1}{c|}{$32.79\pm0.40$} & \multicolumn{1}{c|}{$32.12\pm0.32$} & \multicolumn{1}{c|}{$33.92\pm0.21$} & \multicolumn{1}{c|}{$31.12\pm0.32$} & \multicolumn{1}{c|}{$35.58\pm0.26$} & $\textbf{37.57}\pm\textbf{0.28}$ \\ \cline{3-9} 
 &  & IID & \multicolumn{1}{c|}{$32.61\pm0.21$} & \multicolumn{1}{c|}{$32.19\pm0.10$} & \multicolumn{1}{c|}{$35.32\pm0.17$} & \multicolumn{1}{c|}{$32.20\pm0.12$} & \multicolumn{1}{c|}{$35.44\pm0.19$} & $\textbf{38.26}\pm\textbf{0.19}$ \\ \cline{2-9} 
 & \multirow{4}{*}{ResNet-18} & $\beta=0.1$ & \multicolumn{1}{c|}{$35.12\pm1.00$} & \multicolumn{1}{c|}{$34.63\pm0.82$} & \multicolumn{1}{c|}{$39.06\pm0.59$} & \multicolumn{1}{c|}{$34.82\pm0.82$} & \multicolumn{1}{c|}{$37.18\pm0.35$} & $\textbf{39.28}\pm\textbf{0.58}$ \\ \cline{3-9} 
 &  & $\beta=0.5$ & \multicolumn{1}{c|}{$40.88\pm0.76$} & \multicolumn{1}{c|}{$41.65\pm0.52$} & \multicolumn{1}{c|}{$48.74\pm0.15$} & \multicolumn{1}{c|}{$41.38\pm0.65$} & \multicolumn{1}{c|}{$46.68\pm0.20$} & $\textbf{50.64}\pm\textbf{0.09}$ \\ \cline{3-9} 
 &  & $\beta=1.0$ & \multicolumn{1}{c|}{$43.01\pm0.30$} & \multicolumn{1}{c|}{$43.60\pm0.36$} & \multicolumn{1}{c|}{$46.36\pm0.55$} & \multicolumn{1}{c|}{$42.28\pm0.33$} & \multicolumn{1}{c|}{$47.55\pm0.19$} & $\textbf{51.00}\pm\textbf{0.15}$ \\ \cline{3-9} 
 &  & IID & \multicolumn{1}{c|}{$42.96\pm0.26$} & \multicolumn{1}{c|}{$42.23\pm0.29$} & \multicolumn{1}{c|}{$48.91\pm0.85$} & \multicolumn{1}{c|}{$42.22\pm0.26$} & \multicolumn{1}{c|}{$48.10\pm0.09$} & $\textbf{49.41}\pm\textbf{}\textbf{0.05}$ \\ \cline{2-9} 
 & \multirow{4}{*}{VGG-16} & $\beta=0.1$ & \multicolumn{1}{c|}{$46.77\pm1.24$} & \multicolumn{1}{c|}{$46.38\pm0.85$} & \multicolumn{1}{c|}{$49.81\pm0.81$} & \multicolumn{1}{c|}{$47.28\pm2.55$} & \multicolumn{1}{c|}{$49.23\pm0.65$} & $\textbf{49.99}\pm\textbf{0.70}$ \\ \cline{3-9} 
 &  & $\beta=0.5$ & \multicolumn{1}{c|}{$52.98\pm0.80$} & \multicolumn{1}{c|}{$53.06\pm0.68$} & \multicolumn{1}{c|}{$59.24\pm0.35$} & \multicolumn{1}{c|}{$54.58\pm0.84$} & \multicolumn{1}{c|}{$57.22\pm0.52$} & $\textbf{58.25}\pm\textbf{0.21}$ \\ \cline{3-9} 
 &  & $\beta=1.0$ & \multicolumn{1}{c|}{$56.02\pm0.36$} & \multicolumn{1}{c|}{$55.86\pm0.35$} & \multicolumn{1}{c|}{$58.46\pm0.30$} & \multicolumn{1}{c|}{$55.20\pm0.64$} & \multicolumn{1}{c|}{$57.98\pm0.36$} & $\textbf{59.85}\pm\textbf{0.20}$ \\ \cline{3-9} 
 &  & IID & \multicolumn{1}{c|}{$57.06\pm0.08$} & \multicolumn{1}{c|}{$56.98\pm0.22$} & \multicolumn{1}{c|}{$59.54\pm0.48$} & \multicolumn{1}{c|}{$56.01\pm0.30$} & \multicolumn{1}{c|}{$58.36\pm0.21$} & $\textbf{59.85}\pm\textbf{0.04}$ \\ \hline
\end{tabular}%
}
\vspace{-0.1in}
\end{table*}

\subsection{Experiment Settings}

To assess the effectiveness of our method, we implemented it using the PyTorch framework \cite{pytorch}. For a fair comparison, we utilized the SGD optimizer with a learning rate of 0.01 and a momentum of 0.5. We configured each client's batch size to 50 and conducted 5 epochs for each local training round. In addition, we simulate 100 devices, and 10\% of the devices, i.e. 10 devices, participate in training in each round. Note that in FL, due to considerations of network quality and communication overhead, the server usually only selects some devices to participate in training in each round\cite{review}. All experimental results are from an Ubuntu workstation equipped with an Intel i9 CPU, 64GB memory, and an NVIDIA RTX 4090 GPU. 

\textbf{Datasets and Models. }We selected two well-known datasets in our experiments: the image datasets CIFAR-10 and CIFAR-100~\cite{cifar10}. To verify the effectiveness of our method in IID and non-IID scenarios, we use the Dirichlet distribution Dir($\beta$)\cite{dir} to simulate the data distribution, where the smaller $\beta$ value indicates greater data heterogeneity. To comprehensively evaluate the performance of our method, we conducted experiments on three different models, CNN\cite{cnn}, ResNet-18\cite{resnet}, and VGG-16\cite{vgg}.

\textbf{Baseline. }To show the effectiveness of our proposed method, we carried out experiments using several baseline FL algorithms. The classic FL algorithm FedAvg\cite{fedavg} and four advanced algorithms—CluSamp\cite{cs}, FedNTD\cite{fedntd}, FedGen\cite{fedgen}, and FedMut\cite{hu2024fedmut}—were selected for comparison.
\begin{figure}[h]
\centering
\footnotesize
\subfloat[$\beta = 0.1$]{\includegraphics[width=0.4\columnwidth]{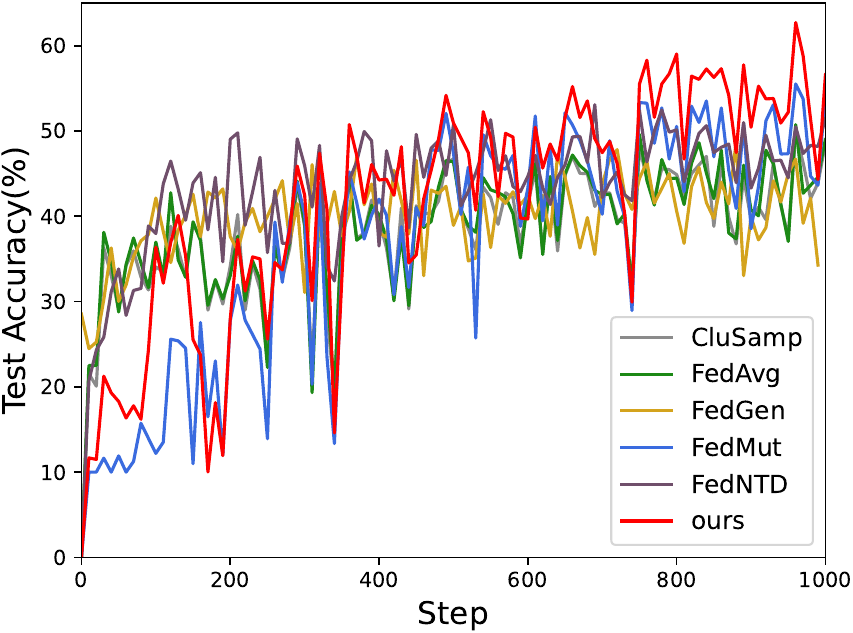}%
\label{}}
\hfil
\subfloat[$\beta = 0.5$]{\includegraphics[width=0.4\columnwidth]{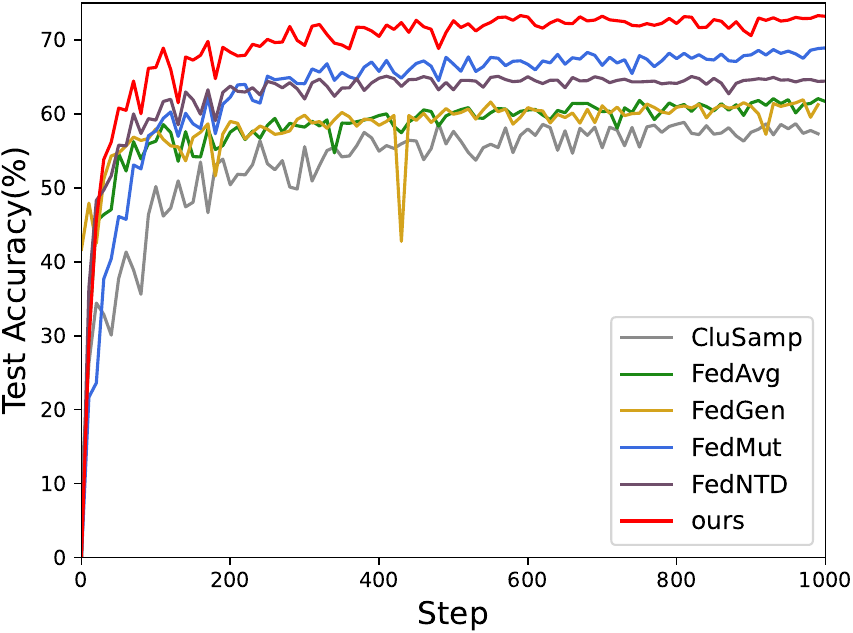}%
\label{}}
\hfil
\subfloat[$\beta = 1.0$]{\includegraphics[width=0.4\columnwidth]{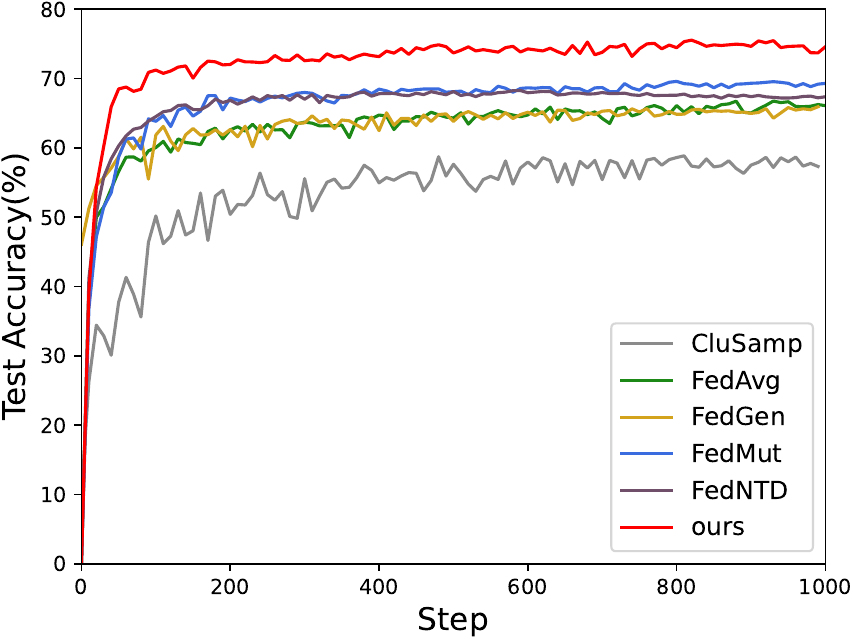}%
\label{}}
\hfil
\subfloat[IID]{\includegraphics[width=0.4\columnwidth]{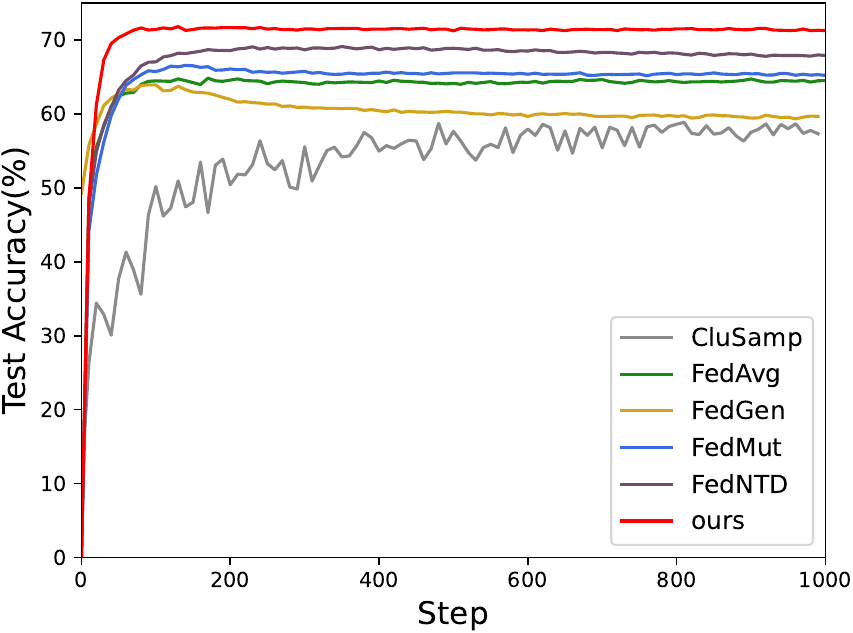}%
\label{}}
\caption{Learning curves of our method and all the baseline methods on CIFAR-10 with ResNet-18.}
\label{fig:acc_compare}
\end{figure}

\subsection{Performance comparison}
\textbf{Comparison of Accuracy. }Table 1 shows the inference accuracy comparison of our method with all baselines under different settings involving different models, datasets, and data distributions. From this table, we can find that our method can always achieve the best performance in all cases.  For example, our method improves test accuracy by 13.68\% over FedAvg in the CIFAR-10 dataset with
VGG-16 model when $\beta=0.1$. Figure \ref{fig:acc_compare} shows the learning curves of our method and all baselines using ResNet-18 and on the CIFAR-10 dataset. We can find that compared with FedMut, our method not only achieves higher accuracy but also converges faster than FedMut in the early stage of training.


\begin{table}[h]
\centering
\caption{Comparison of test accuracy under different numbers of participating clients.}
\label{tab:diff-frac}
\resizebox{\linewidth}{!}{%
\begin{tabular}{|cc|c|c|c|}
\hline
\multicolumn{2}{|c|}{Values of $K$} & 5 & 10 & 20 \\ \hline
\multicolumn{1}{|c|}{\multirow{6}{*}{\begin{tabular}[c]{@{}c@{}}Test \\ Accuracy (\%)\end{tabular}}} & FedAvg & $59.15\pm0.56$ & $61.72\pm0.58$ & $62.01\pm0.45$ \\ \cline{2-5} 
\multicolumn{1}{|c|}{} & CluSamp & $60.14\pm0.32$ & $61.35\pm0.48$ & $62.58\pm0.66$ \\ \cline{2-5} 
\multicolumn{1}{|c|}{} & FedNTD & $65.76\pm0.65$ & $64.96\pm0.24$ & $66.42\pm0.24$ \\ \cline{2-5} 
\multicolumn{1}{|c|}{} & FedGen & $61.45\pm0.28$ & $60.92\pm0.69$ & $58.43\pm0.79$ \\ \cline{2-5} 
\multicolumn{1}{|c|}{} & FedMut & $70.11\pm0.46$ & $68.60\pm0.16$ & $65.25\pm0.22$ \\ \cline{2-5} 
\multicolumn{1}{|c|}{} & Ours & $\textbf{73.93}\pm\textbf{0.29}$ & $\textbf{74.09}\pm\textbf{0.15}$ & $\textbf{70.21}\pm\textbf{0.28}$ \\ \hline
\end{tabular}%
}
\end{table}

\textbf{Impacts of Different Numbers of Participating Clients. } In our previous experiments, we assumed that 10\% of the devices participated in training in each round. To assess how varying the number of clients participating in training affects the accuracy of our method, we designed two different experimental scenarios. Specifically, we select 5 and 20 devices to participate in training in each round, i.e., $K=5,20$, respectively. We conducted the experiment using the ResNet-18 model and the CIFAR-10 dataset with $\beta=0.5$. Table \ref{tab:diff-frac} shows the experimental results. We can find that our method can achieve the highest accuracy, which can prove the effectiveness of our method in different AIoT scenarios.


\subsection{Ablation Study}

\begin{figure}[h]
    \centering
    \begin{subfigure}[b]{0.25\textwidth}
        \includegraphics[width=\textwidth]{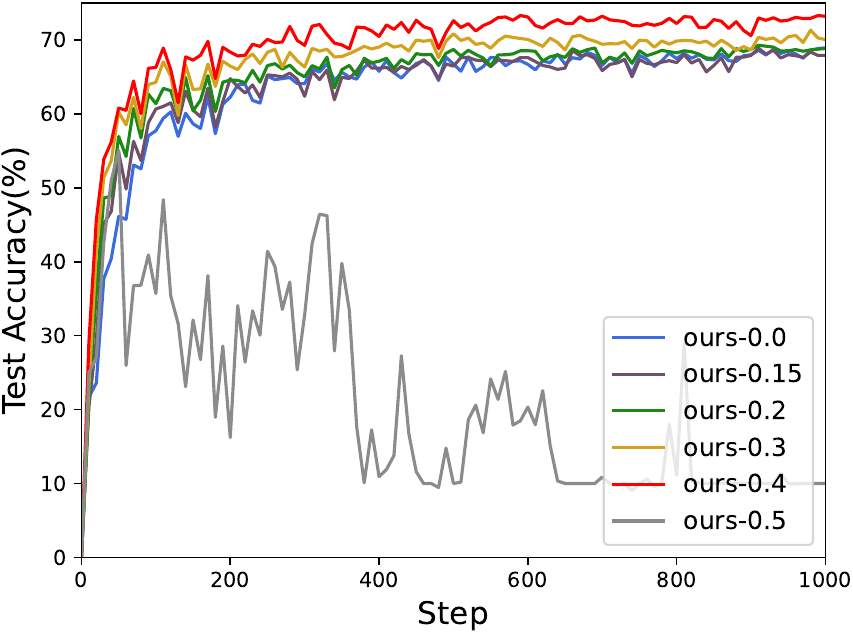}
    \end{subfigure}
    \caption{Ablation study results.}
    \label{fig:Ablation}
    \vspace{-0.1 in}
\end{figure}

To explore the specific impact of quadratic programming rate on the performance of our method, we conducted experiments using the ResNet-18 model and CIFAR-10 dataset with $\beta=0.5$. Figure \ref{fig:Ablation} shows the learning curves of FedMut and our method under different settings. The label ``ours-p" represents our method with \(p\) as the dynamic programming probability. In particular, when \( p = 0 \), our method is the same as FedMut.

Our experimental results show that our method exhibits significant improvements in accuracy as the quadratic programming probability \( p \) increases. This shows that appropriately increasing the value of \( p \) can effectively optimize model performance. However, further testing also found that when the value of \( p \) is too high, the stability of the model begins to be affected, manifested as a slowdown in convergence or a complete inability to reach the convergence state. This phenomenon may be because the high-probability quadratic programming causes the model to search too frequently in the parameter space, thereby preventing the model from stabilizing near the optimal solution. Therefore, although increasing \( p \) can improve the learning ability of the model, the upper limit of \( p \) also needs to be carefully controlled to avoid negative effects caused by over-adjustment.

\subsection{Discussion}
\textbf{Limitations.} Our approach relies on a carefully tuned quadratic programming probability to achieve higher accuracy. The optimal quadratic programming probability varies across different models and datasets, so the generalization ability and robustness of our method are limited.

\textbf{Future Work.} 
Based on the experimental results of this paper, we plan to explore the specific reasons why the model fails to converge with a high quadratic programming probability \( p \). We will identify which factors most affect the stability and performance of the model through detailed parameter sensitivity analysis and loss landscape visualization. In addition, we expect to adopt an adaptive adjustment mechanism, combined with real-time feedback, to continuously adjust the value of \( p \) through experiments, aiming to identify the value \( p \) that achieves the optimal balance between model performance and convergence. 
In addition, this paper only focuses on the inference accuracy of the global model, without considering various uncertain factors in physical environments~\cite{hu2020quantitative,chen2024flexfl} and fairness problems in training data~\cite{li2023fairer, li2023fairness}.
In the future, we will attempt to improve FedQP to adapt to various uncertain environments and address fairness problems.


\section{Conclusion}
This paper proposed a novel mutation-based FL approach named FedQP, aiming to address the lack of directionality when exploring the parameter space. With a quadratic programming-guided mutation region design, our approach can make the global model mutate towards the fan-shaped area of the gradient. This design enhances the directional control of the mutation process, allowing the mutation to proceed in a targeted direction guided by the gradient, thus enabling the model to explore specific directions more strategically. The experimental results exhibit the effectiveness of our proposed method on different datasets, indicating that the quadratic programming-guided mutation strategy can improve the performance of FL.

\section{Acknowledgement}
This work was supported by Natural Science Foundation of China (62272170),   and ``Digital Silk Road'' Shanghai International Joint Lab of Trustworthy Intelligent Software (22510750100), Shanghai Trusted Industry Internet Software Collaborative Innovation Center, the Startup Foundation for Introducing Talent of NUIST (No.
2024r035), and the Automatic Software Generation and Intelligent Service Key Laboratory of Sichuan Province (No. CUIT-SAG202306)

\bibliographystyle{IEEEtran}
\bibliography{reference}

\end{document}